%% file: main.tex
\documentclass[conference]{IEEEtran}
\usepackage{times}

\usepackage[numbers]{natbib}
\usepackage{multicol}
\usepackage[bookmarks=true]{hyperref}
\usepackage{lipsum}
\usepackage{CJK}
\usepackage{amsmath}
\usepackage{algorithm}
\usepackage{algpseudocode}
\usepackage{graphicx}
\usepackage{xspace}
\usepackage{multirow}
\usepackage{booktabs}
\usepackage{subcaption}
\usepackage{tabularx}
\usepackage{array}
\usepackage[dvipsnames]{xcolor}
\usepackage{listings}

\usepackage{color, colortbl}
\definecolor{lightblue}{HTML}{e6f3ff}
\definecolor{lightorange}{HTML}{fff2e6}
\definecolor{lightgreen}{HTML}{90EE90}
\definecolor{darkblue}{HTML}{0065a2}
\definecolor{darkorange}{HTML}{ffa23a}

\newcommand{\model}{\textsc{ViLa}}

\makeatletter
\def\blfootnote{\xdef\@thefnmark{}\@footnotetext}
\makeatother

\begin{document}

\title{Look Before You Leap: Unveiling the Power of GPT-4V in Robotic Vision-Language Planning}

\author{Yingdong Hu$^{1,2,3*}$, \quad
Fanqi Lin$^{1,2,3*}$, \quad
Tong Zhang$^{1,2,3}$, \quad
Li Yi$^{1,2,3}$, \quad
Yang Gao$^{1,2,3\dagger}$ \\
$^{1}$Tsinghua University \hspace{0.2in}
$^{2}$Shanghai Artificial Intelligence Laboratory  \hspace{0.2in}
$^{3}$Shanghai Qi Zhi Institute \\
\{huyd21, lfq20, zhangton20\}@mails.tsinghua.edu.cn, \{ericyi, gaoyangiiis\}@mail.tsinghua.edu.cn\\
\href{https://robot-vila.github.io/}{robot-vila.github.io}
}

\twocolumn[{%
\renewcommand\twocolumn[1][]{#1}%
\maketitle
\begin{center}
    \centering 
    \includegraphics[width=\linewidth]{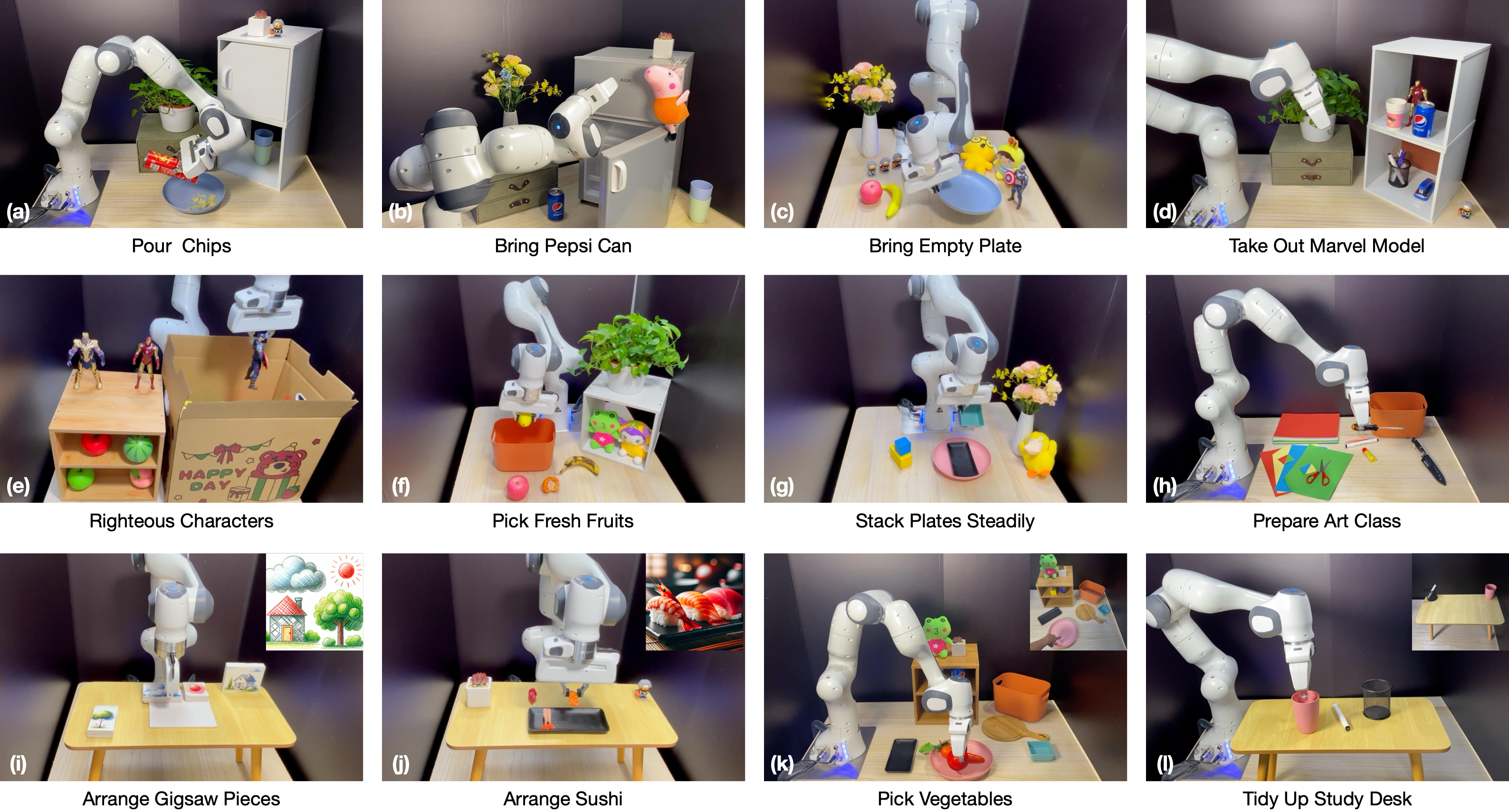}
    \captionof{figure}{We present \textbf{\model}, a simple and effective method for long-horizon robotic task planning. By integrating vision directly into the reasoning process, \model~can leverage the wealth of commonsense knowledge grounded in the visual world. This results in remarkable performance in tasks that demand an understanding of spatial layouts (top row), object attributes (middle row), and tasks with multimodal goals (bottom row).}
    \label{fig:teaser}
\end{center}
}]
{\blfootnote{{$^{*}$ The first two authors contributed equally.}}}
{\blfootnote{{$^{\dagger}$ Correspondence to: Yang Gao \textless{}gaoyangiiis@tsinghua.edu.cn\textgreater{}.}}}

\begin{abstract}
In this study, we are interested in imbuing robots with the capability of physically-grounded task planning.
Recent advancements have shown that large language models (LLMs) possess extensive knowledge useful in robotic tasks, especially in reasoning and planning. 
However, LLMs are constrained by their lack of world grounding and dependence on external affordance models to perceive environmental information, which cannot jointly reason with LLMs.
We argue that a task planner should be an inherently grounded, unified multimodal system.
To this end, we introduce Robotic Vision-Language Planning (\model), a novel approach for long-horizon robotic planning that leverages vision-language models (VLMs) to generate a  sequence of actionable steps.
\model~directly integrates perceptual data into its reasoning and planning process, enabling a profound understanding of commonsense knowledge in the visual world, including spatial layouts and object attributes.
It also supports flexible multimodal goal specification and naturally incorporates visual feedback.
Our extensive evaluation, conducted in both real-robot and simulated environments, demonstrates \model's superiority over existing LLM-based planners, highlighting its effectiveness in a wide array of open-world manipulation tasks.
\end{abstract}

\IEEEpeerreviewmaketitle

\input{documents/intro}
\input{documents/related}
\input{documents/method}
\input{documents/exp}
\input{documents/conclusion}

\section*{Acknowledgments}
This work is supported by the Ministry of Science and Technology of the People's Republic of China, the 2030 Innovation Megaprojects ``Program on New Generation Artificial Intelligence'' (Grant No. 2021AAA0150000).
This work is also supported by the National Key R\&D Program of China (2022ZD0161700).


\bibliographystyle{plainnat}
\bibliography{references}

\input{documents/appendix}

\end{document}

%% file: documents/intro.tex
\section{Introduction}

Scene-aware task planning is a pivotal facet of human intelligence~\cite{wilensky1983planning,suchman1987plans}. When presented with a simple language instruction, humans demonstrate a spectrum of complex behaviors depending on the context. Take the instruction ``get a can of coke,'' for example. If a coke can is visible, a person will immediately pick it up. If not, they will search locations like the refrigerator or storage cabinets. This adaptability reflects humans' deep understanding of the scene and extensive common sense, enabling them to interpret instructions contextually. In this paper, we explore how we can create an embodied agent, such as a robot, that emulates this human-like adaptability and exhibits long-horizon task planning in varying scenes. 

In recent years, large language models (LLMs)~\cite{brown2020language,openai2023gpt4,chowdhery2022palm,bommasani2021opportunities} have showcased their remarkable capabilities in encoding extensive semantic knowledge about the world~\cite{petroni2019language,jiang2020can,gurnee2023language}. This has sparked a growing interest in leveraging LLMs for generating step-by-step plans for complex, long-horizon tasks~\cite{ahn2022can,huang2022language,huang2022inner}. However, a critical limitation of LLMs is their lack of world grounding — they cannot perceive and reason about the physical state of robots and their environments, including object shapes, physical properties, and real-world constraints.

To overcome this challenge, a prevalent approach involves employing external affordance models~\cite{gibson1977theory}, such as open-vocab detectors~\cite{minderer2205simple} and value functions~\cite{ahn2022can}, to provide real-world grounding for LLMs~\cite{ahn2022can, huang2023grounded}. However, these modules often fail to convey the truly necessary task-dependent information in complex environments, as they serve as one-directional channels transmitting perceptual information to LLMs. In this scenario, the LLM is like a blind person, while the affordance model serves as a sighted guide. On the one hand, the blind person relies solely on their imagination and the guide's limited narrative to comprehend the world; on the other hand, the sighted guide may not accurately comprehend the blind person's purpose. This combination often leads to unfeasible or unsafe action plans in the absence of precise, task-relevant visual information. For instance, a robot tasked with taking out a Marvel model from a shelf (see Figure~\ref{fig:teaser} (d)) may overlook obstacles like the paper cup and coke can, leading to collisions. Consider another example of preparing art class (see Figure~\ref{fig:teaser} (h)), scissors can be perceived as sharp and hazardous objects, or as essential tools for handicrafts. This distinction is challenging for the vision module due to the lack of specific task information. These examples highlight the limitations of LLM-based planners in capturing intricate spatial layouts and fine-grained object attributes, underscoring the necessity for active joint reasoning between vision and language.

The recent advancements in vision-language models (VLMs), exemplified by GPT-4V(ision)~\cite{gpt4v, yang2023dawn}, have significantly broadened the horizons of research. VLMs synergize perception and language processing into a unified system, enabling direct incorporation of perceptual information into the language model’s reasoning~\cite{liu2023visual,instructblip,bai2023qwen,zhu2023minigpt}. Building upon these developments, we introduce Robotic \textbf{Vi}sion-\textbf{La}nguage Planning (\textbf{\textsc{ViLa}}) — a simple, effective, and scalable method for long-horizon robotic planning. \model~distinguishes itself from previous LLM-based planning methods by eschewing independent affordance models and instead directly prompting VLMs to generate a sequence of actionable steps based on visual observations of the environment and high-level language instructions. \model~exhibits the following key properties absent in LLM-based planning methods:

\begin{itemize}
     \item \textbf{Profound Understanding of Commonsense Knowledge Grounded in the Visual World.} \model~excels in complex tasks that demand an understanding of spatial layouts (e.g., \texttt{Take Out Marvel Model}) or object attributes (e.g., \texttt{Stack Plates Steadily}). This kind of commonsense knowledge pervades nearly every task of interest in robotics, but previous LLM-based planners consistently fall short in this regard.
    \item \textbf{Versatile Goal Specifiaction.} \model~supports flexible multimodal goal specification approaches. It is capable of utilizing not just language instructions but also diverse forms of goal images, and even a blend of both language and images, to define objectives effectively.
    \item \textbf{Visual Feedback.} \model~effectively utilizes visual feedback in an intuitive and natural way, enabling robust closed-loop planning in dynamic environments.
\end{itemize}

We conduct a systematic evaluation of \model~across 16 real-world, everyday manipulation tasks, which involve a diverse range of open-set instructions and objects. \model~consistently outperforms LLM-based planners, such as SayCan~\cite{ahn2022can} and Grounded Decoding~\cite{huang2023grounded}, by a significant margin. To facilitate a more exhaustive and rigorous comparison, we extend our evaluation to include 16 simulated tasks based on the RAVENS environment~\cite{zeng2021transporter}, wherein \model~continues to show marked enhancements. All these outcomes provide compelling evidence that \model~possesses the potential to serve as a universal task planning method for general-purpose robotic systems.

%% file: documents/related.tex
\section{Related Work}

\noindent \textbf{Vision-Language Models.}
The striking advancements made by scaling up large language models (LLMs)~\cite{brown2020language,openai2023gpt4,chowdhery2022palm,touvron2023llama,hoffmann2022training} have sparked a surge of interest in similarly expanding large vision-language models (VLMs)~\cite{gan2022vision,gpt4v,yang2023dawn}. 
The prevalent approach to construct VLMs involves employing a cross-modal connector to align the features of pre-trained visual encoders with the input embedding space of the LLMs~\cite{alayrac2022flamingo,liu2023visual,liu2023improved,chen2023shikra,li2023blip,huang2023language,ye2023mplug,zhu2023minigpt,bai2023qwen,wang2023cogvlm}. 
The ability of VLMs to understand both images and text renders them highly adaptable for a range of applications, including visual question answering~\cite{antol2015vqa,zellers2021merlot}, image captioning~\cite{agrawal2019nocaps,hu2022scaling}, and optical character recognition~\cite{li2023trocr}.
In contrast to these uses, our study takes a different path. We concentrate on harnessing the rich world knowledge and the visually grounded attribute of VLMs to address complex long-horizon planning challenges in robotics.

\smallskip
\noindent \textbf{Pre-Trained Foundation Models For Robotics.} Recent advancements in applying large pre-trained foundation models to robotics can be classified into three categories:

(1) Pre-Trained Vision Models: A wealth of prior approaches employ vision models pre-trained on large-scale image datasets~\cite{grauman2022ego4d,deng2009imagenet} to generate visual representations for visuomotor control tasks~\cite{parisi2022unsurprising,xiao2022masked,radosavovic2023real,nair2022r3m,ma2022vip,zhang2023universal}. Nonetheless, a robotic system encompasses more than just a perception module; it includes a control policy as well. Relying solely on visual representations that capture high-level semantics may not ensure the control policy's generalizability or the system's overall effectiveness~\cite{hu2023pre,yuan2023rl,hansen2022pre}.

(2) Pre-Trained Language Models: Another research avenue explores the use of large language models (LLMs) for robotic tasks, particularly in reasoning and planning~\cite{huang2022language,ahn2022can,singh2023progprompt,wu2023tidybot,song2023llm,liu2023llm+,vemprala2023chatgpt,lin2023text2motion,ding2023task,ren2023robots}. However, to ground these language models in physical environments, auxiliary modules such as affordance models~\cite{huang2023grounded}, perception APIs~\cite{liang2023code}, and textual scene descriptions~\cite{huang2022inner,zeng2022socratic} are essential. In contrast, our work emphasizes generating plans without depending on these auxiliary models for grounding. This approach allows for the seamless integration of perceptual information directly into the reasoning and planning process.

(3) Pre-Trained Vision-Language Models: Numerous studies have explored the application of vision-language models (VLMs) in robotics~\cite{huang2023voxposer,du2023vision,shah2023lm,zhang2023grounding,gao2023physically}. Notably, RT-2~\cite{brohan2023rt} demonstrates the integration of VLMs in low-level robotic control. In contrast, our research is primarily centered on high-level robotic planning. Although PaLM-E~\cite{driess2023palm} shares similarities with our approach, it necessitates training on a substantial mixture of robotics and general visual-language data~\cite{chen2022pali,lynch2023interactive}. This approach implies that introducing a robot to a new environment necessitates the collection of additional data and subsequent retraining of the model. In stark contrast, our \model~stands out as an open-world, zero-shot model. It is capable of performing a broad spectrum of everyday manipulation tasks without additional training data and in-context examples in the prompt.

\begin{figure*}[t!]
    \centering
    \includegraphics[width=1.0\linewidth]{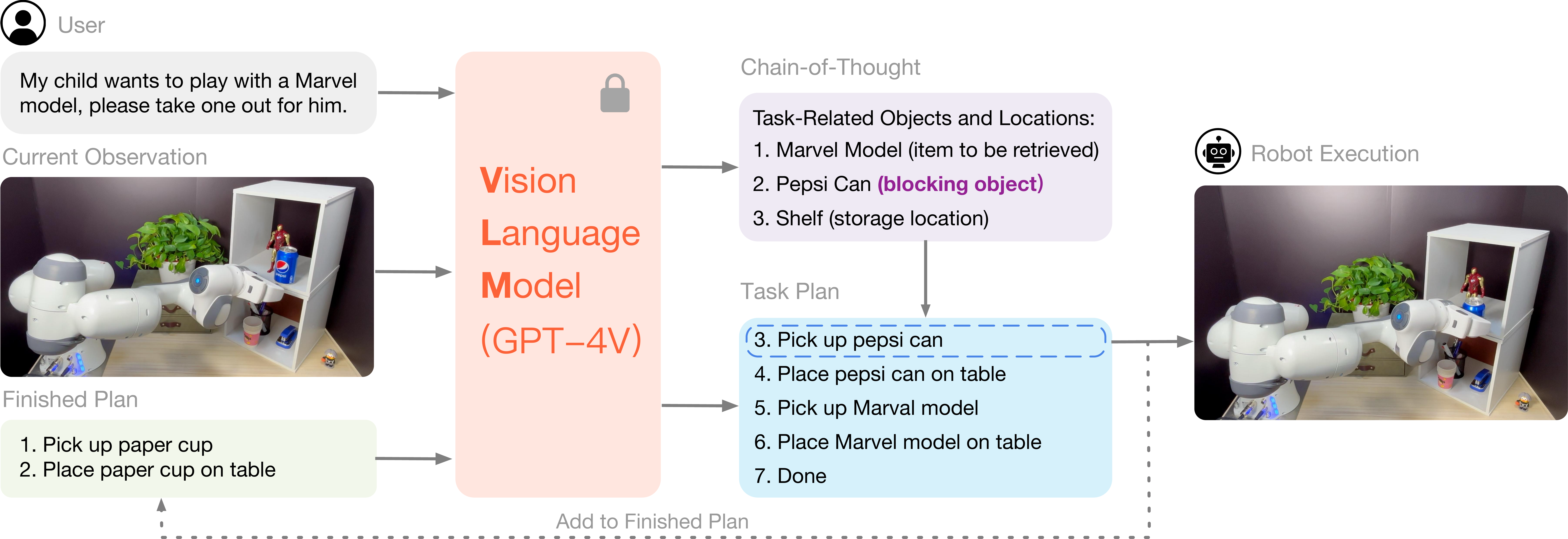}
    \caption{\textbf{Overview of \model.} Given a language instruction and current visual observation, we leverage a VLM to comprehend the environment scene through chain-of-though reasoning, subsequently generating a step-by-step plan. The first step of this plan is then executed by a primitive policy. Finally, the step that has been executed is added to the finished plan, enabling a closed-loop planning method in dynamic environments.}
    \label{fig:method}
\end{figure*}

\smallskip
\noindent \textbf{Task and Motion Planning.} Task and Motion Planning (TAMP)~\cite{kaelbling2011hierarchical,garrett2021integrated} stands as a critical framework in solving long-horizon planning tasks, integrating low-level continuous motion planning~\cite{lavalle2006planning} with high-level discrete task planning~\cite{fikes1971strips,sacerdoti1975structure,nau1999shop}. While traditional research in this domain has predominantly centered on symbolic planning~\cite{fikes1971strips,nau1999shop} or optimization-based methods~\cite{toussaint2015logic,toussaint2018differentiable}, the advent of machine learning~\cite{xu2018neural,eysenbach2019search,garrett2020online,ichter2020broadly} and LLMs~\cite{ding2023task,chen2023autotamp,silver2023generalized,xie2023translating} is revolutionizing this arena. In our work, we leverage VLMs to comprehend the robot environment and interpret high-level instructions. By incorporating commonsense knowledge that is intrinsically grounded in the visual world, our approach excels in handling complex tasks beyond the reach of previous LLM-based planning methods.

%% file: documents/method.tex
\section{Method}

We first provide the formulation of the planning problem in Sec.~\ref{sec:problem}. Subsequently, we present how \model~utilize vision-language models as robot planners (Sec.~\ref{sec:robot-plan}). Finally, we describe unique properties of \model~that contribute to its advantages (Sec.~\ref{sec:property}).

\subsection{Problem Statement}
\label{sec:problem}
Our robotic system takes a visual observation $\mathbf{x}_t$ of the environment and a high-level language instruction $\mathcal{L}$ (e.g. ``\textit{stack these containers of different colors steadily}'') that describes a manipulation task. We assume that the visual observation $\mathbf{x}_t$ serves as an accurate representation of world state. The language instruction $\mathcal{L}$ can be arbitrarily long-horizon or under-specified (i.e., requires contextual understanding). The central problem investigated in this work is to generate a sequence of text actions, represented as $\ell_1, \ell_2, \cdots, \ell_T$. Each text action $\ell_t$ is a short-horizon language instruction (e.g. ``\textit{pick up blue container}'') that specifies a sub-task/primitive skill $\pi_{\ell_t} \in \Pi$. Note that our contributions do not focus on the acquisition of these skills $\Pi$; rather, we assume that all the necessary skills are already available. These skills can take the form of predefined script policies or may have been acquired through various learning methods, including reinforcement learning (RL)~\cite{sutton2018reinforcement} and behavior cloning (BC)~\cite{pomerleau1988alvinn}.

\begin{algorithm}[t]
\caption{~~\model}\label{alg:main}
\begin{algorithmic}[1]
\Require Initial visual observation $\mathbf{x}_1$, a high level instruction $\mathcal{L}$ and a set of skills $\Pi$.
\State $t = 1$, $\ell_1=\emptyset$
\While{$\ell_{t-1} \neq \text{``done''}$}
\State $p_{1:N} = \operatorname{VLM}(\mathbf{x}_t, \mathcal{L}, \ell_1, ..., \ell_{t-1})$ \Comment{Get plan steps}
\State $\ell_t = p_{1}$ \Comment{Select the first step}
\State Execute skill $\pi_{\ell_{t}}(\mathbf{x}_t)$, updating observation $\mathbf{x}_{t+1}$
\State $t = t + 1$
\EndWhile
\end{algorithmic}
\end{algorithm}

\subsection{Vision-Language Models as Robot Planners}
\label{sec:robot-plan}

To generate feasible plans, high-level robot planning must be grounded in the physical world. While LLMs possess a wealth of structured world knowledge, their exclusive reliance on language input necessitates external components, such as affordance models, to complete the grounding process. However, these external affordance models (e.g., value functions of RL policies~\cite{ahn2022can,kalashnikov2021mt}, object detection models~\cite{minderer2205simple}, and action detection models~\cite{shridhar2021cliport}) are manually designed as independent channels, operating separately from LLMs, rather than being integrated into an end-to-end system. Moreover, their role is solely transmitting high-dimensional visual perceptual information to LLMs, lacking the capability for joint reasoning. This separation of vision and language modalities results in the vision module's inability to provide comprehensive, task-relevant visual information, thereby hindering the LLM from planning based on accurate task-related visual insights.

Recent advances in vision-language models (VLMs) offer a solution. VLMs demonstrate unprecedented ability in understanding and reasoning across both images and language~\cite{liu2023visual,instructblip,bai2023qwen,zhu2023minigpt}. Crucially, the extensive world knowledge encapsulated in VLMs is inherently grounded in the visual data they process. Therefore, we advocate for directly employing VLMs that synergizes vision and language capabilities to decompose a high-level instruction into a sequence of low-level skills. 

We refer to our method as Robotic \textbf{Vi}sion-\textbf{La}nguage Planning (\textbf{\textsc{ViLa}}). Concretely, given current visual observation $\mathbf{x}_t$ of environment and a high-level language goal $\mathcal{L}$, \model~operates by prompting the VLMs to yield a step-by-step plan $p_{1:N}$. We enable closed-loop execution by selecting the first step as the text action $\ell_t = p_1$. Once the text action $\ell_t$ is selected, the corresponding policy $\pi_{\ell_t}$ is executed by the robot and the VLM query is amended to include $\ell_t$ and the process is run again until a termination token (e.g., ``done'') is reached. The entire process is shown in Figure~\ref{fig:method} and described in Algorithm~\ref{alg:main}.

In our study, we utilizes GPT-4V(ision)~\cite{gpt4v, yang2023dawn} as the VLM. 
GPT-4V, trained on vast internet-scale data, exhibits exceptional versatilities and extremely strong generalization capabilities. These attributes make it particularly adept at handling open-world scenarios presented in our paper.
Furthermore, we find that \model, powered by GPT-4V, is capable of solving a variety of challenging planning problems, even when operating in a \textit{zero-shot} mode (i.e., without requiring any in-context examples). This significantly reduces the prompt engineering efforts required in previous approaches~\cite{ahn2022can,huang2022language,huang2023grounded}.

\subsection{Intriguing Properties of \model}
\label{sec:property}
In this section, we delve deeper into \model, shedding light on its advantages and differentiations from previous planning methods.

\smallskip
\noindent \textbf{Comprehension of Common Sense in the Visual World.}
Previous studies primarily focus on leveraging the knowledge of LLMs for high-level planning~\cite{ahn2022can,huang2022language}, centering on language while often overlooking the crucial role of vision.
Images and languages, as distinct types of signals, offer unique nature: languages are human-generated and semantically rich, yet they are limited in their ability to represent comprehensive information. In contrast, images are natural signals imbued with low-level fine-grained features, a single image can capture the entirety of a scene's information. 
This disparity is especially pertinent when the complex environment is challenging to encapsulate in simple language. Directly integrating images into the reasoning and planning process, such as in the case of \model, allows for a more intuitive understanding of commonsense knowledge grounded in the physical world. Specifically, this understanding manifests in two key aspects:

\subsubsection{Spatial Layout Understanding}
Describing complex geometric configurations, particularly spatial localization, object relationships, and environmental constraints, can be challenging with just simple language. 
Consider a cluttered scene where object A obscures object B. To reach object B, one must first reposition object A. Relying solely on verbal language descriptions to convey these nuanced relationships between objects is inadequate. 
Moreover, consider a situation where the desired object is inside a container (like a cabinet or refrigerator). In that case, if an external affordance model (like object detection model) is utilized, since the desired object is not visible, the affordance model would predict a zero probability of successful retrieval, leading to task failure. 
However, by directly incorporating vision into the reasoning process, \model~can deduce that the sought object, hidden from view, is likely inside the container. This realization necessitates opening the container as a preliminary step to accomplish the task.

\subsubsection{Object Attribute Understanding}
An object is defined by multiple attributes, including its shape, color, material, function, etc. However, the expressive capacity of natural language is limited, making it a somewhat cumbersome medium for conveying these attributes comprehensively. 
Furthermore, note that an object's attributes is intricately tied to the specific tasks at hand. 
For example, scissors might be deemed hazardous for children, but they become essential tools during a paper-cutting art class.
Previous approaches employ a standalone affordance model to identify object attributes, but this method can only convey a limited subset of attributes in a unidirectional manner. Therefore, active joint reasoning between image and language emerges as a crucial necessity when our tasks demand a thorough understanding of an object's attributes.

\begin{figure*}[t!]
    \centering
    \includegraphics[width=1.0\linewidth]{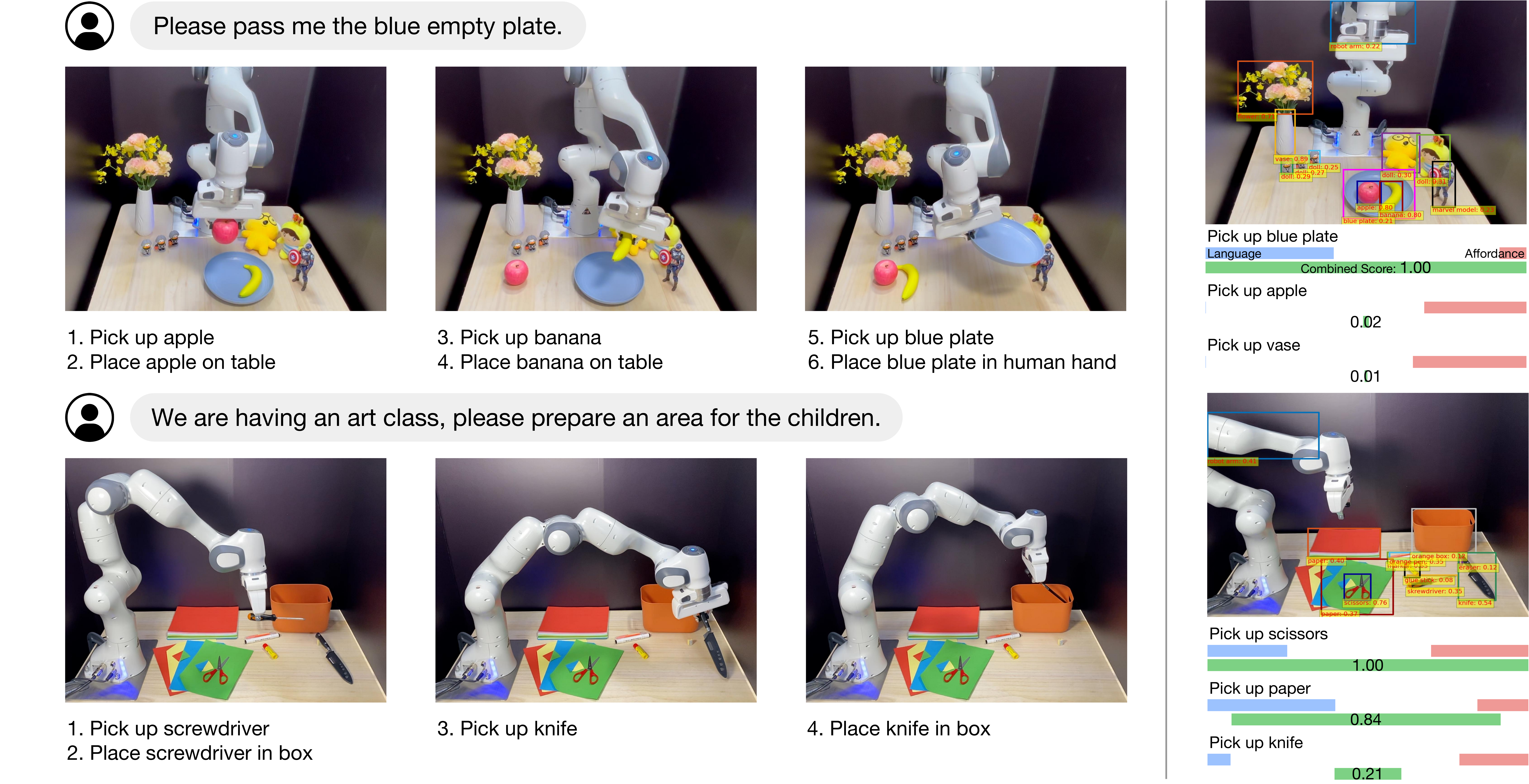}
    \caption{\textbf{Illustration of the execution of \model~(left) and the decision-making process of SayCan (right)}. In the \texttt{Bring Empty Plate} task, the robot must first relocate the apple and banana from the blue plate. However, SayCan's initial step is to directly pick up the blue plate. In the \texttt{Prepare Art Class} task, while the scissor is supposed to remain on the table, SayCan erroneously picks up the scissor and places it in a box.}
    \label{fig:rollouts}
\end{figure*}

\smallskip
\noindent \textbf{Versatile Goal Specification.}
In many complex, long-term tasks, using a goal image to represent the desired outcome is often more effective than relying solely on verbal instructions. 
For example, to direct a robot to tidy a desk, providing a photo of the desk arranged as desired can be more efficient. Likewise, for food plating tasks, a robot can replicate the arrangement from an image. Such tasks, previously unattainable with LLM-based planning methods, are now remarkably straightforward with \model. 
Specifically, \model~can not only accepts current visual observation $\mathbf{x}_n$ and language instructions $\mathcal{L}$ as inputs but also incorporates a goal image $\mathbf{x}_g$. 
This feature sets it apart from many existing goal-conditioned RL/IL algorithms ~\cite{nair2018visual,eysenbach2022contrastive,ding2019goal}, as it does not require the goal and visual observation images to originate from the same domain. 
The goal image merely needs to convey the essential elements of the task, offering flexibility in its form – it could range from an internet photo to a child's drawing, or even an image showing a target location indicated by a pointing finger. This versatility greatly enhances the system's practicality. Additionally, the ability to combine images and language in describing task goals introduces an additional layer of flexibility and diversity in our goal specification approach.

\smallskip
\noindent \textbf{Visual Feedback.}
The embodied environments are inherently dynamic, making closed-loop feedback essential for robots. In an effort to incorporate environment feedback into planning methods that rely solely on LLMs, Huang et al.~\cite{huang2022inner} investigate converting all feedback to natural language. However, this approach proves to be cumbersome and ineffective because most of the feedback is initially observed visually. Converting visual feedback into language not only adds complexity to the system but also risks losing valuable information. We believe that providing visual feedback directly is a more intuitive and natural approach, as demonstrated in \model. Within \model, the VLM serves both as a scene descriptor to recognize object states and as a success detector to determine if the environment satisfies the success conditions defined by the instructions. By reasoning over visual feedback, \model~enables robots to make corrections or replan in response to changes in the environment or when a skill fails.

%% file: documents/exp.tex
\section{Experiments and Analysis}

In this section, we first carry out extensive experiments in a real-world system to evaluate \model's capability in planning everyday manipulation tasks (Sec.~\ref{sec:real-world}). Subsequently, we conduct a detailed quantitative comparison of \model~against baseline methods within a simulated tabletop environment (Sec.~\ref{sec:simulation}).

\subsection{Real-World Manipulation Tasks}
\label{sec:real-world}

\noindent \textbf{Experimental Setup.}

\subsubsection{Hardware} We set up a real-world tabletop environment. We use a Franka Emika Panda robot (a 7-DoF arm) and a 1-DoF parallel jaw gripper. For perception, we use a Logitech Brio color camera mounted on a tripod, at an angle, pointing towards the tabletop. To ensure consistency in our experiments, we maintain a fixed camera view for all tasks, but for visual aesthetics, we record video demos at different views.

\subsubsection{Tasks and Evaluation} We design 16 long-horizon manipulation tasks to assess \model's performance in three domains: 
comprehension of commonsense knowledge in the visual world (8 tasks), flexibility in goal specification (4 tasks), and utilization of visual feedback (4 tasks). Figure~\ref{fig:teaser} illustrates a selection of 12 tasks, drawn from the first two domains. 
For each task, we evaluate all methods across the 10 different variations of the environment, including changes in scene configuration and lighting conditions, etc. For comprehensive details of each task, please see Appendix~\ref{sec:tasks_and_evaluation}.

\subsubsection{VLM and Prompting} We use GPT-4V from~\href{https://openai.com/api/}{OpenAI API} as our VLM. Unlike previous approaches~\cite{ahn2022can,huang2023grounded}, we do not include any in-context examples in the prompt, but only use high-level language instructions and some simple constraints that the robot needs to meet (i.e., strict \textit{zero-shot}). The full prompt is shown in Appendix~\ref{sec:real-world-prompt}.

\subsubsection{Primitive Skills}
We use five categories of primitive skills that lend themselves to complex behaviors through composition and planning. These include ``pick up \texttt{object}'', ``place \texttt{object} in/on \texttt{object}'', ``open \texttt{object}'', ``close \texttt{object}'', and ``pour \texttt{object} into/onto \texttt{object}''. We concentrate on high-level, temporally extended planning rather than acquiring low-level primitive skills, which is orthogonal to our study. Therefore, we employ script policies as the primitive skills for both the baselines and \model. Additional details of low-level primitive skills are in Appendix~\ref{sec:primitive-skill}.

\begin{table}[t]
    \normalsize
    \centering
    \begin{tabular}{lccc}
         \toprule
         \textbf{Task} & \textbf{SayCan} & \textbf{GD} & \textbf{\model} \\
         \midrule
         \textcolor{darkblue}{Pour Chips}         & 20\% & 40\% & 80\%              \\
         \textcolor{darkblue}{Bring Pepsi Can}           & 40\% & 30\% & 90\%        \\
         \textcolor{darkblue}{Bring Empty Plate}          & 0\%  & 0\%  & 100\%      \\
         \textcolor{darkblue}{Take Out Marvel Model}      & 0\%  & 10\% & 70\%       \\
         \midrule
         \textcolor{darkorange}{Righteous Characters}           & 0\% & 10\% & 80\%  \\
         \textcolor{darkorange}{Pick Fresh Fruits}              & 20\% & 30\% & 80\% \\
         \textcolor{darkorange}{Stack Plates Steadily}          & 20\% & 10\% & 70\% \\
         \textcolor{darkorange}{Prepare Art Class}              & 0\% & 30\% & 70\%  \\
         \midrule
         \textbf{Total}                       & 13\% & 20\% & 80\% \\  
         \bottomrule
    \end{tabular}
    \caption{\textbf{Quantitative evaluation results in tasks requiring rich commonsense knowledge.} \model~demonstrates superior performance in tasks necessitating a understanding of \textcolor{darkblue}{spatial layouts} (top half) and \textcolor{darkorange}{object attributes} (bottom half).}
    \label{tab:commonsense}
\end{table}

\subsubsection{Baselines} 
We compare with SayCan~\cite{ahn2022can} and Grounded Decoding (GD)~\cite{huang2023grounded}, which both ground LLMs with external affordance models.
Implementing these baselines necessitates accessing output token probabilities from LLMs. However, since ~\href{https://openai.com/api/}{OpenAI API} currently does not return these probabilities, we employ the open-source Llama 2 70B~\cite{touvron2023llama} as an alternative.
For the affordance models, we utilize the open-vocabulary detector OWL-ViT~\cite{minderer2205simple,minderer2023scaling}, following Huang et al~\cite{huang2023grounded}.

\smallskip
\noindent \textbf{\model~can understand commonsense knowledge in the visual world.} In Table~\ref{tab:commonsense}, we compare the planning success rates on tasks that require understanding of spatial layouts and object attributes (see Figure~\ref{fig:teaser} (a-h) for illustrations of the tasks). \model~stands out with an average success rate of 80\% across 8 tasks, significantly surpassing the performances of SayCan and GD, which achieve success rates of only 13\% and 20\%, respectively. Particularly in intricate and challenging tasks such as \texttt{Take Out Marvel Model}~(it's crucial to avoid the cup and coke can) and \texttt{Righteous Characters}~\footnote{Choose righteous characters from three Marvel models, while referring to the model only by its color. Details of this task are in Appendix~\ref{sec:tasks_and_evaluation}.}, SayCan and GD's success rates are close to \textit{zero}. These tasks all necessitate the integration of images into the reasoning and planning processes and a deep understanding of commonsense knowledge in the visual world. Furthermore, the tasks outlined in Table~\ref{tab:commonsense} are representative of typical real-world scenarios and are not specifically tailored for \model. The across-the-board exceptional performance of \model~not only highlights its superior generalizability but also underscores its potential as a universal planner for open-world tasks.

\begin{figure}[t]
    \centering
    \includegraphics[width=1.0\linewidth]{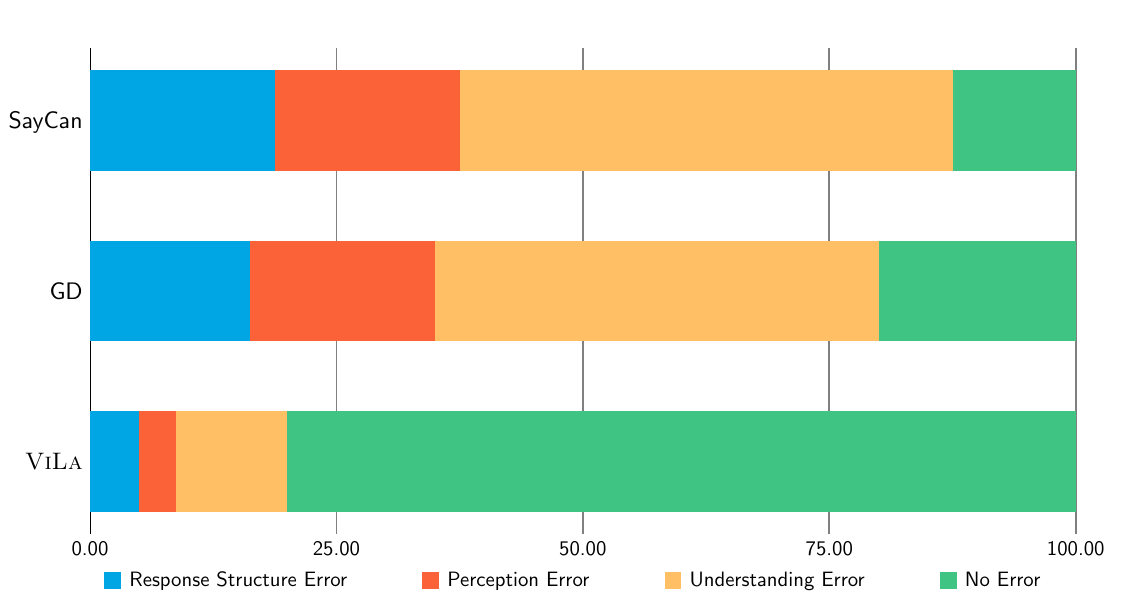}
    \caption{\textbf{Error breakdown of \model~and baselines.} By leveraging commonsense knowledge grounded in the visual world, \model~significantly reduces understanding error.}
    \label{fig:error_breakdown}
\end{figure}

Figure~\ref{fig:rollouts} shows two environment rollouts comparing \model~with SayCan. In the first \texttt{Bring Empty Plate} task, \model~identifies the need to relocate the apple and banana from the blue plate before picking it up. In contrast, SayCan recognizes the items (apple, banana, blue plate) but lacks awareness of their spatial relationship, leading it to attempt picking up the blue plate directly. This highlights the significance of comprehending complex geometric configurations and environmental constraints visually. In another scenario involving the preparation of a safe area for a children's art class (\texttt{Prepare Art Class}), \model~discerns that only the screwdriver and fruit knife are hazardous, sparing the scissors necessary for the class, based on the contextual clue of paper cuttings on the table. However, SayCan misclassifies the scissors as dangerous, showing that a comprehensive, global visual understanding is crucial to accurately assess object attributes. The videos of experiment rollouts can be found on the project website: \href{https://robot-vila.github.io/}{robot-vila.github.io}.

\begin{figure*}[t]
    \centering
    \includegraphics[width=1.0\linewidth]{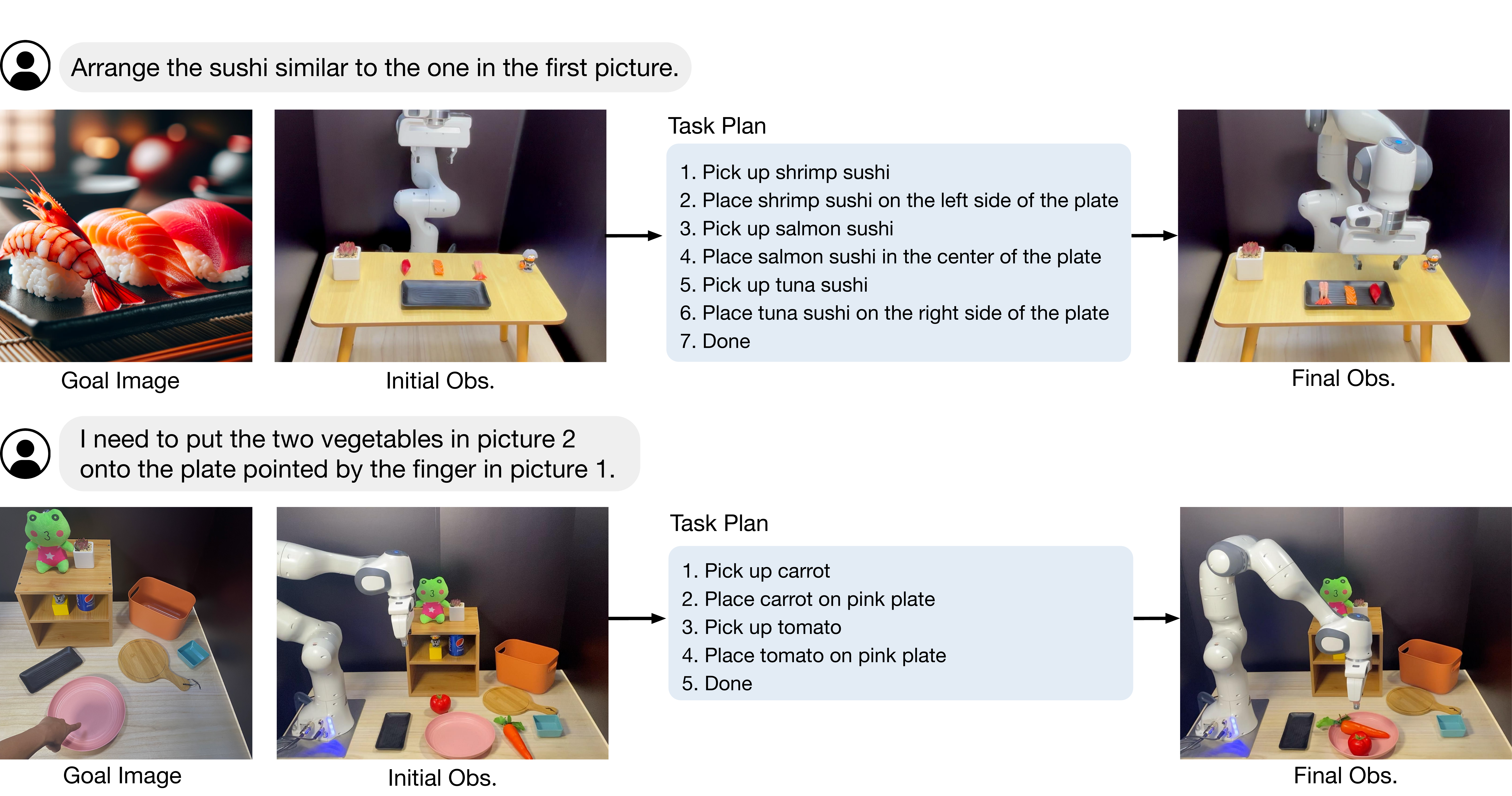}
    \caption{\textbf{Illustration of the execution of \model~on image goal-conditioned tasks.} In the \texttt{Arrange Sushi} task, \model~generates a plan to arrange sushi based on a reference image. In the \texttt{Pick Vegetables} task, the scenario involves a table set with a pink plate, a black sushi plate, a pizza plate, and a green snack plate. Here, \model~deduces from pointing finger in the goal image that the vegetables should be placed on the pink plate.}
    \label{fig:goal_condition}
\end{figure*}

In Figure~\ref{fig:error_breakdown}, we present a failure breakdown analysis. ``Response structure error'' here refers to errors of LLMs and VLMs in generating plan steps that fall outside our predefined set of primitive skills. In the case of baselines, ``perception error'' denotes failures within the open-vocab detector~\cite{minderer2023scaling}. While VLMs lack a separate perception module, their output, as observed in the chain-of-thought process~\cite{wei2022chain}, occasionally fails to recognize some objects. The dominant error in baseline models is ``understanding error'', which involves errors in understanding the complex spatial layouts and object attributes in the physical world, such as occlusions and context-specific attributes. \model~significantly reduces the ``understanding error'' by seamlessly integrating vision and language reasoning, thereby resulting in the lowest overall error. Furthermore, we suggest that careful prompt engineering (i.e., providing examples in the prompt)~\cite{brown2020language,ouyang2022training} could steer VLM outputs towards admissible primitive skills, thereby reducing ``response structure error''.

\begin{table}[t]
    \normalsize
    \centering
    \begin{tabular}{lcc}
         \toprule
         \textbf{Task} & \textbf{Goal Type} & \textbf{Succ. \%} \\
         \midrule 
         Arrange Sushi                    & Real Image & 80\%           \\
         Arrange Gigsaw Pieces            & Drawing &   100\%           \\
         Pick Vegetables                  & Pointing Finger & 100\%     \\
         Tidy Up Study Desk               & Image + Language & 60\%     \\
         \bottomrule
    \end{tabular}
    \caption{\textbf{Quantitative evaluation results of \model~in tasks featuring multimodal goals.} }
    \label{tab:goal}
\end{table}

\smallskip
\noindent \textbf{\model~supports flexible multimodal goal specification.} We introduce a suite of 4 tasks, each with distinct goal types, as illustrated in Figure~\ref{fig:teaser} (i-l). The quantitative results are shown in Table~\ref{tab:goal}, where \model~demonstrates strong capabilities across all tasks. Utilizing the internet-scale knowledge imbued in GPT-4V, \model~exhibits the remarkable ability to understand a variety of goal images. This includes interpreting vibrant children's drawings for puzzle completion, preparing a sushi platter by referencing a photograph of the dish (illustrated in Figure~\ref{fig:goal_condition} top row), and even accurately identifying the intended arrangement of vegetables as indicated by a human finger (refer to Figure~\ref{fig:goal_condition} bottom row). Additionally, we explore goal specification through a combination of image and language instructions. For instance, in the \texttt{Tidy Up Study Desk} task, we not only provide an image of a neatly organized desk as the target but also verbally direct the swapping of two specific objects on the desk. Leveraging its dual-capacity in vision and language reasoning, \model~consistently achieves success in this task as well. 

\begin{table}[t]
    \normalsize
    \centering
    \begin{tabular}{lcc}
         \toprule
         \textbf{Task} & \textbf{Open-Loop} & \textbf{w/ Feedback} \\
         \midrule
         Stack Blocks                           & 20\% & 90\%  \\
         Pack Chip Bags                         & 0\%  & 100\% \\
         Find Stapler                           & 30\% & 90\% \\
         Human-Robot Interaction                & 20\% & 80\% \\
         \bottomrule
    \end{tabular}
    \caption{\textbf{Open-loop \model~vs. closed-loop \model.} By leveraging visual feedback, closed-loop \model~substantially outperforms the open-loop variant.}
    \label{tab:feedback}
\end{table}

\begin{figure*}[t!]
    \centering
    \includegraphics[width=1.0\linewidth]{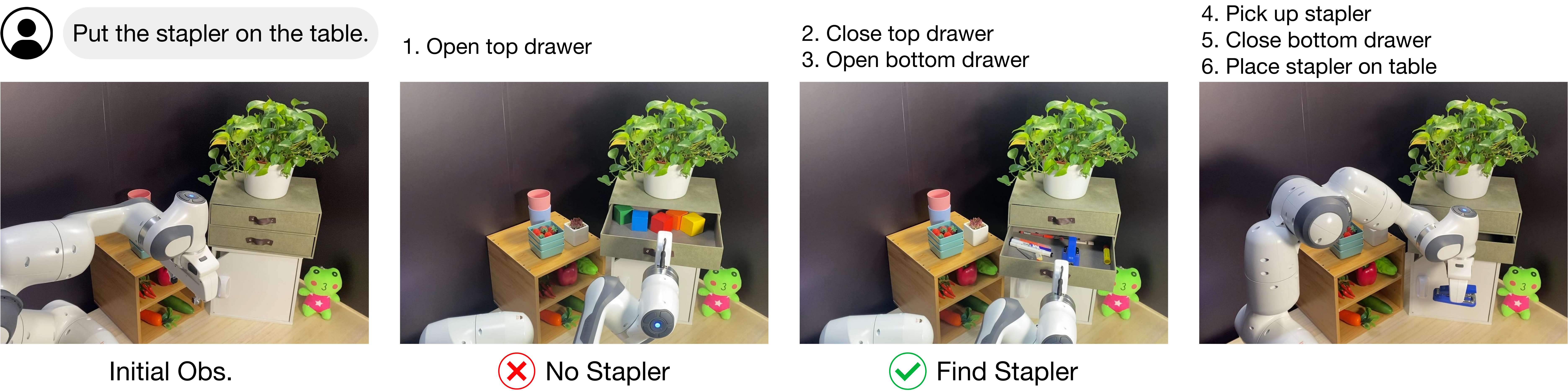}
    \caption{\textbf{Illustration of the execution of \model~on the \texttt{Find Stapler} task.} By incorporating visual feedback and replanning at every step, \model~is able to continue exploring the bottom drawer when it does not find the stapler in the top drawer, thereby successfully locating the stapler.}
    \label{fig:feedback}
\end{figure*}

\begin{figure}[t!]
    \centering
    \includegraphics[width=1.0\linewidth]{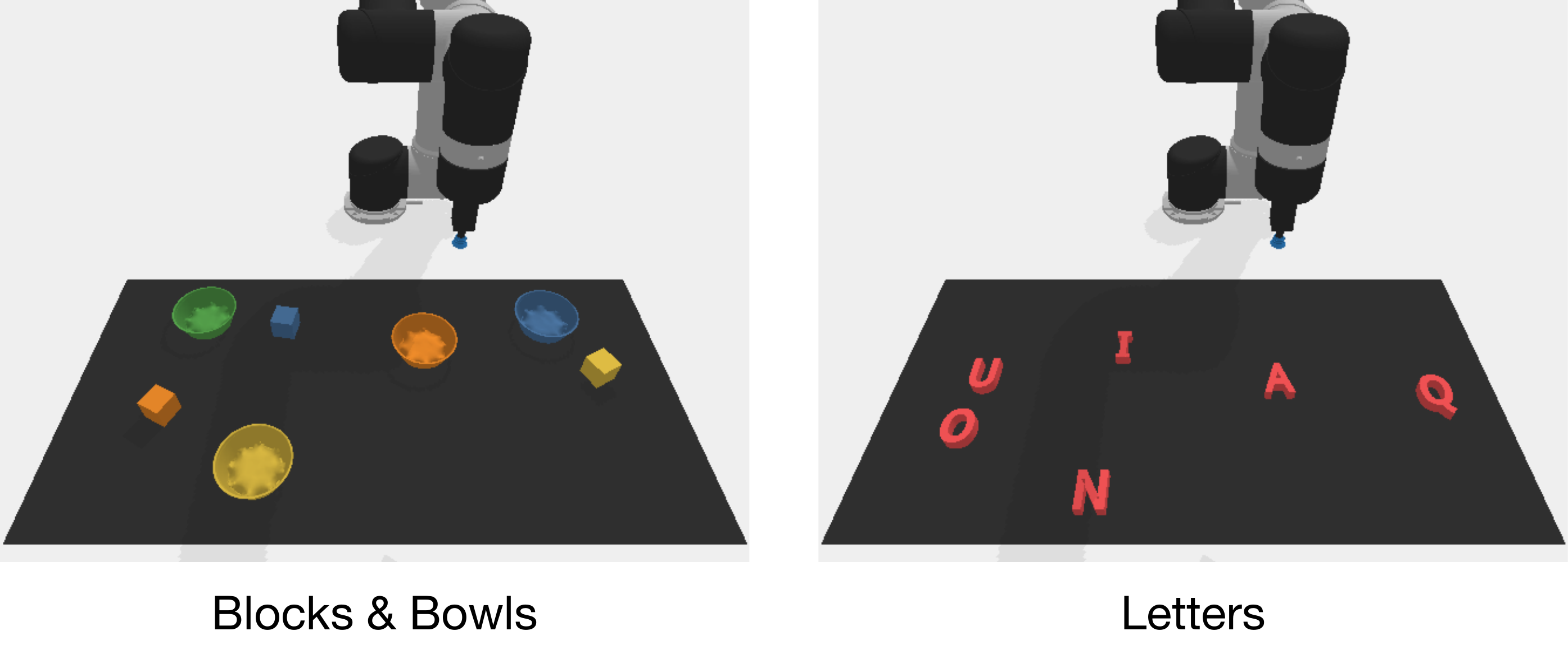}
    \caption{\textbf{Simulated environment based on RAVENS.} We design 16 distinct tasks, which are grouped into two categories: \texttt{Blocks \& Bowls} (left) and \texttt{Letters} (right).}
    \label{fig:simulation}
\end{figure}

\smallskip
\noindent \textbf{\model~can leverage visual feedback naturally.}
We design 4 tasks that require real-time visual feedback for successful execution. In the \texttt{Stack Blocks} task, we inject Gaussian noise into the joint position controller, which increases the likelihood of failure in the primitive policy. 
For the \texttt{Pack Chip Bags} task, task progress is reverted by an experimenter who takes out previously packed chip bags from the box. 
In the \texttt{Find Stapler} task, the stapler's location varies among three potential places: the top drawer, the bottom drawer, or the cabinet. 
The \texttt{Human-Robot Interaction} task requires the robot to pause until a person retrieves the cola it has picked up. 
We evaluate the performance of \model~against an open-loop variant that formulates a plan based solely on the initial observation.
The quantitative results, presented in Table~\ref{tab:feedback}, reveal that the open-loop variant struggles with these dynamic tasks that demand continuous replanning, while the closed-loop \model~significantly outperforms it. 
\model~is not only able to effectively recover from external disturbances but can also adapt its strategy based on real-time visual observations.
A case in point, depicted in Figure~\ref{fig:feedback}, is when \model, not finding the stapler in the top drawer, proceeds to check the bottom drawer, successfully locates the stapler, and completes the task.

\subsection{Simulated Tabletop Rearrangement}
\label{sec:simulation}

\noindent \textbf{Experimental Setup.} 
We conduct experiments on simulated tabletop rearrangement tasks to provide a more rigorous and fair comparison with baseline methods. Following the setting in Grounded Decoding~\cite{huang2023grounded}, we develop 16 tasks based on the RAVENS environment~\cite{zeng2021transporter}. These tasks are categorized into two groups: a seen group, consisting of 6 tasks used for few-shot prompting or as training for supervised baselines, and an unseen group of 10 tasks. Each task requires a UR5 robot to rearrange the objects on the table in some desired configuration, specified by high-level language instructions. The tasks are further classified into two types (see Figure~\ref{fig:simulation}): \textit{(i)} \texttt{Blocks \& Bowls} (8 tasks), which focus on rearranging or combining blocks and bowls (e.g., ``put all the blocks in the bowls with matching colors''). \textit{(ii)} \texttt{Letters} (8 tasks), which involve rearranging alphabetical letters (e.g., ``put the letters on the tables in alphabetical order''). More details about the environmental setup are in Appendix~\ref{sec:app-simulation}

\begin{table}[t]
\begin{center}
\setlength\tabcolsep{2.0pt}
\small
\bgroup
\def\arraystretch{1.4}
\begin{tabular}{@{}lcccccc@{}}
    \toprule
      \multicolumn{1}{l}{} & \multicolumn{2}{c}{\textbf{CLIPort}} & \multicolumn{2}{c}{\textbf{LLM}} & \multirow{2}{*}{\textbf{GD}}  & \multirow{2}{*}{\textbf{\model}} \\
      \cmidrule(lr){2-3} \cmidrule(lr){4-5} 
    \textbf{Tasks} & Short & Long & Llama 2 & GPT-4 \\

    \midrule
    \normalsize{\textbf{\textcolor{gray}{Seen Tasks}}}\\
    \small{Blocks \& Bowls}   & 3.3\% & 68.3\% & 1.7\% & 0\% & 18.3\% & 78.3\% \\
    \small{Letters}   & 0\% & 40.0\% & 25.0\% & 25.0\% & 51.7\% & 88.3\% \\
    
    \midrule
    \normalsize{\textbf{\textcolor{gray}{Unseen Tasks}}} \\
    \small{Blocks \& Bowls}   & 6.0\% & 6.0\% & 20.0\% & 22.0\% & 23.0\% & 81.0\% \\
    \small{Letters}   & 1.0\% & 0\% & 15.0\% & 15.0\% & 42.0\% & 82.0\% \\
    \bottomrule
\end{tabular}
\egroup
\caption{\textbf{Average success rate in simulated environment.} See Appendix~\ref{sec:sim_full_results} for a detailed breakdown. \model~consistently outperforms baselines across seen and unseen tasks.}
\label{table:ravens_simple}
\end{center}
\end{table}

Our comparison encompasses three baseline categories: \textit{(i)} CLIPort~\cite{shridhar2021cliport}, a language-conditioned imitation learning agent that directly take in the high-level language instructions without a planner. We consider two variants: ``Short'', trained on single-step pick-and-place instructions, and ``Long'', trained on high-level instructions. \textit{(ii)} An LLM-based planner that does not relay on any grounding/affordance model. We evaluate Llama 2 and GPT-4. \textit{(iii)} Grounded Decoding (GD), which integrates an LLM with an affordance model for enhanced planning. Here, Llama 2 is used as the LLM. For tasks in \texttt{Blocks \& Bowls}, affordances are derived from CLIPort's predicted logits, while for tasks in \texttt{Letters}, we use ground-truth affordance values obtained from simulation. We use script policies as the primitive skills for LLM-based planner, GD and our \model.

\smallskip
\noindent \textbf{Analysis.}
The results are presented in Table~\ref{table:ravens_simple}, where each method is evaluated over 20 episodes per task within each category.
We observe that CLIPort-based methods have a limited capacity for generalizing to novel, unseen tasks. Given that GD requires access to the output token probabilities of LLMs, we employ Llama 2 instead of GPT-4 for GD. As depicted in Table~\ref{table:ravens_simple}, both Llama 2 and GPT-4 exhibit comparable performances across all tasks, ensuring a fair comparison between GD and \model~(utilizing GPT-4V). While GD surpasses other LLM-based planning methods by leveraging an external affordance model, it significantly lags behind \model. This finding further highlights the benefits of synergistic reasoning between vision and language for high-level robotic planning.

%% file: documents/conclusion.tex
\section{Conclusion, Limitations, \& Future Works} 
\label{sec:conclusion}

In this work, we present \model, a novel approach for robotic planning that utilizes VLMs to decompose a high-level language instruction into a sequence of actionable steps. \model~integrates perceptual information into the reasoning and planning process, enabling the understanding of commonsense knowledge in the visual world (e.g., spatial layouts and object attributes). It also supports flexible multimodal goal specification and naturally integrates visual feedback. Our extensive evaluation, conducted in both real-world and simulated settings, demonstrate \model's effectiveness in addressing a variety of complex, long-horizon tasks.

\model~has several limitations that future work can improve. First, we presuppose the existence of all single-step primitive skills. While obtaining robust low-level control policies remains a challenging problem, recent advancements in transferring web knowledge to robotic control~\cite{brohan2022rt,brohan2023rt} holds promise for enabling the cultivation of a repertoire of generalizable skills. Secondly, our dependence on a black-box VLM hampers steerability and complicates the explanation of certain errors. Future developments could leverage parameter-efficient fine-tuning methods~\cite{houlsby2019parameter,hu2021lora} to customize VLMs~\cite{gao2023physically}. Finally, our current approach excludes in-context examples within prompts, leading to a more versatile output format. Methods developed for prompting~\cite{kojima2022large,yao2023tree} can also be used to refine output consistency.

%% file: documents/appendix.tex
\clearpage
\onecolumn
\appendix 
\counterwithin{figure}{subsection}
\counterwithin{table}{subsection}

\subsection{Real-World Environment}
\label{sec:app-real-world}

\subsubsection{Hardware Setup}
We use a Franka Emika Panda robot (a 7-DoF arm) and a 1DoF parallel jaw gripper. The robot is operated using the joint controller from Deoxys~\cite{zhu2022viola}. For perception, we use a Logitech Brio color camera mounted on a tripod, at an angle, pointing towards the tabletop. This camera offers high-resolution images at 1920 × 1080, ensuring maximum detail retention. 

\subsubsection{Tasks and Evaluation}
\label{sec:tasks_and_evaluation}
We design 16 long-horizon tasks, categorized into three domains: \textit{(i)} understanding commonsense knowledge in the visual world (8 tasks, detailed in Table~\ref{tab:commonsense_tasks}); \textit{(ii)} flexibility in goal specification (4 tasks, detailed in Table~\ref{tab:goal_tasks}); and \textit{(iii)} utilization of visual feedback (4 tasks, detailed in Table~\ref{tab:feedback_tasks}). For each task, we perform 10 evaluations under different variations of the environment, accounting for changes in scene configuration, lighting conditions, etc.

\subsubsection{Prompts}
\label{sec:real-world-prompt}
\begin{figure*}[h!]
    \centering
    \includegraphics[width=1.0\linewidth]{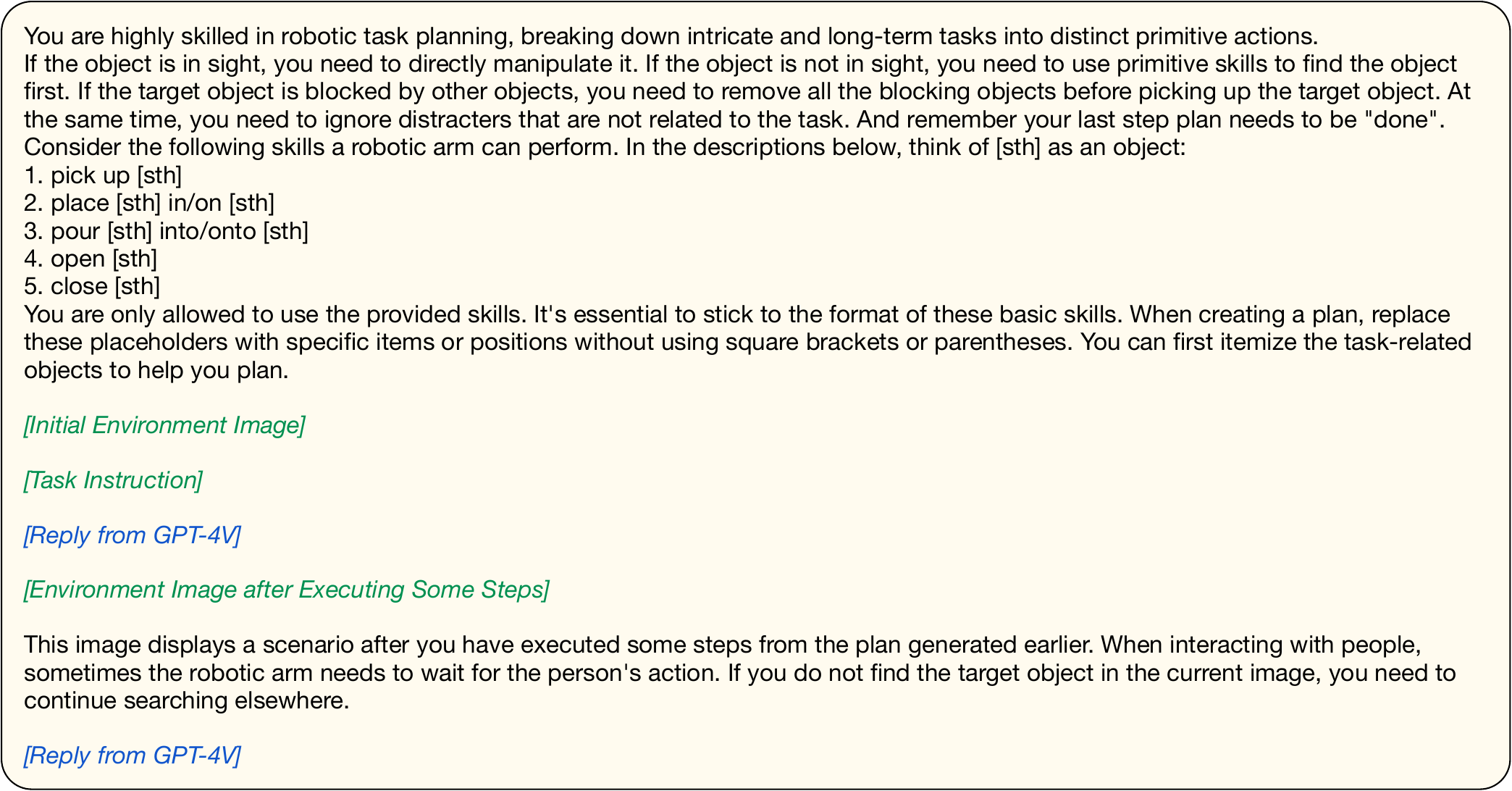}
    \caption{\textbf{Prompt for real-world environment.}}
    \label{fig:real_world_prompt}
\end{figure*}

We do not include any in-context examples in the prompt, but only use high-level language instructions and some simple constraints that the robot needs to meet (i.e., strict zero-shot). The full prompt is shown in Figure~\ref{fig:real_world_prompt}.

\subsubsection{Primitive Skills}
\label{sec:primitive-skill}
We use five categories of primitive skills that lend themselves to complex behaviors through composition and planning. These include ``pick up \texttt{object}'', ``place \texttt{object} in/on \texttt{object}'', ``open \texttt{object}'', ``close \texttt{object}'', and ``pour \texttt{object} into/onto \texttt{object}''. We concentrate on high-level, temporally extended planning rather than acquiring low-level primitive skills, which is orthogonal to our study. Therefore, we employ script policies as the primitive skills. For simple tasks like ``pick up \texttt{object}'', we teleoperate the robots by operating a 3D SpaceMouse. For more intricate, contact-rich tasks such as ``open drawer'', kinematic teaching is employed. These skills are tailored to the tasks considered in our study,
developing a generalizable and robust set of primitive skills is an important area for future exploration and research.

\begin{table*}[h!]
    \small
    \centering
    \begin{tabular}{m{14em}m{40em}}

    \toprule
    \textcolor{darkblue}{\texttt{Pour Chips}}
    & \textbf{Instruction:} ``My child is hungry, please pour him a plate of chips.'' \\
    & \textbf{Description:} In five of the evaluation episodes, the chips are stored inside a cabinet, requiring the robot arm to first open the cabinet in order to locate the chips. For the remaining five episodes, the chips are directly visible, the robot arm should immediately pick up the chip bag. \\
    \noalign{\vskip 1mm}
    \hline
    \noalign{\vskip 1mm}
    \textcolor{darkblue}{\texttt{Bring Pepsi Can}}  & \textbf{Instruction:} ``I'm very thirsty, can you help me get a can of cola and put it on the table?'' \\
    & \textbf{Description:} In five of the evaluation episodes, the Pepsi can is placed inside a refrigerator, requiring the robot arm to first open the refrigerator to locate the Pepsi can. In the other five episodes, the Pepsi can is directly visible, the robot arm should immediately grasp and pick it up. \\
    
        \noalign{\vskip 1mm}
    \hline
    \noalign{\vskip 1mm}
    \textcolor{darkblue}{\texttt{Bring Empty Plate}} & \textbf{Instruction:} ``Please pass me the blue empty plate.'' \\
    & \textbf{Description:} In each evaluation episode, one or two objects are placed on a blue plate, surrounded by several distractor objects on the table. The robot arm is required to first remove the objects on the plate and then hand the plate to a human, while disregarding the distractor objects. \\
    
        \noalign{\vskip 1mm}
    \hline
    \noalign{\vskip 1mm}
    \textcolor{darkblue}{\texttt{Take Out Marvel Model}}
     & \textbf{Instruction:} ``My child wants to play with a Marvel model, please take one out for him.'' \\
    & \textbf{Description:} The Marvel model is placed on the upper shelf, blocked by one or two objects, with additional distractor objects placed on the lower shelf, on top of the shelf, or around the shelf. The robot arm is required to first remove the objects blocking the Marvel model, then pick up the model and place it on the table, while ignoring the distractor objects. \\
    
        \noalign{\vskip 1mm}
    \hline
    \noalign{\vskip 1mm}
    \textcolor{darkorange}{\texttt{Righteous Characters}} & \textbf{Instruction:} ``I want to pick some righteous character models for my child, but I am not familiar with these characters. Which color toys should I put in the box?'' \\
    & \textbf{Description:}  There are three Marvel character models: Iron Man (righteous), Captain America (righteous), and Thanos (unrighteous). Due to the constraints in the instruction stating, ``I am not familiar with these characters'' and ``which color toys should I ...'', the robot plan must not explicitly mention the names of the characters (such as Iron Man), but is limited to referencing models by their color (like red model). \\

        \noalign{\vskip 1mm}
    \hline
    \noalign{\vskip 1mm}
    \textcolor{darkorange}{\texttt{Pick Fresh Fruits}} & \textbf{Instruction:} ``I want to buy some fruits. Help me pick the fresh fruits from this pile of fruits and put them into the orange box.'' \\
    & \textbf{Description:}  There are some rotten fruits and some incomplete fruits (such as a half-peeled orange). The robot arm needs to disregard these distracting fruits and accurately select the fresh fruits. \\

        \noalign{\vskip 1mm}
    \hline
    \noalign{\vskip 1mm}
    \textcolor{darkorange}{\texttt{Stack Plates Steadily}} & \textbf{Instruction:} ``Steadily stack these containers of different colors.'' \\
    & \textbf{Description:}  There are several containers of varying sizes and colors. The robot arm must accurately discern the relative sizes of these containers and stack them steadily in order of size. \\

        \noalign{\vskip 1mm}
    \hline
    \noalign{\vskip 1mm}
    \textcolor{darkorange}{\texttt{Prepare Art Class}} & \textbf{Instruction:} ``We are having an art class, please prepare an area for the children. Please put any inappropriate items on the table into the box.'' \\
    & \textbf{Description:}  Certain objects are unsuitable for an art class setting (such as screwdrivers and fruit knives), while others (like glue and colored paper) are appropriate. Classifying scissors is challenging as they can be viewed as either hazardous or a craft tool for cutting paper. In this specific context, with paper cuttings present, scissors should be retained for this task. This task requires the task planner to ground objects within the specific scene to determine their attributes. \\
    
    \bottomrule
    
    \end{tabular}
    \caption{\textbf{A list of 8 tasks requiring understanding commonsense knowledge in the visual world.} The first four tasks are centered on comprehending \textcolor{darkblue}{spatial layouts}, while the subsequent four are dedicated to understanding \textcolor{darkorange}{object attributes}. For every task, we provide the instruction as used in our experiments and a detailed description of the task.}
    \label{tab:commonsense_tasks}
\end{table*}

\begin{table*}[h]
    \small
    \centering
    \begin{tabular}{m{14em}m{40em}}

    \toprule

    \texttt{Arrange Sushi} & \textbf{Instruction:} ``In the second picture, arrange the sushi on a specific side of the plate similar to the one in the first picture.'' \\
    & \textbf{Goal Type}: Real Image \\
    & \textbf{Description:}  The planner needs to identify the types of sushi and their arrangement in the goal image, and then, based on the observed image from the experiment, place the sushi onto a specific location on the sushi plate. \\

        \noalign{\vskip 1mm}
    \hline
    \noalign{\vskip 1mm}
    \texttt{Arrange Jigsaw Pieces} & \textbf{Instruction:} ``The first picture is my child’s drawing. In the second picture, arrange the jigsaw pieces on the corners of the whiteboard similar to the landscape image shown in the first picture.'' \\
    & \textbf{Goal Type}: Drawing \\
    & \textbf{Description:} The planner needs to identify the positions of elements in the goal image, and then, based on the observed image from the experiment, place the jigsaw pieces in a specific corner of the whiteboard. \\

        \noalign{\vskip 1mm}
    \hline
    \noalign{\vskip 1mm}
    \texttt{Pick Vegetables} & \textbf{Instruction:} ``I need to put the two vegetables in picture 2 onto the plate pointed by the finger in picture 1.'' \\
    & \textbf{Goal Type}: Pointing Finger \\
    & \textbf{Description:} In the goal image, there are multiple plates of different colors. The planner is required to identify the plate being pointed at by a finger, and based on the image observed in the experiment, place the vegetables from the scene onto the indicated plate. \\

        \noalign{\vskip 1mm}
    \hline
    \noalign{\vskip 1mm}
    \texttt{Tidy Up Study Desk} & \textbf{Instruction:} ``Study the arrangement in the first picture. Replicate it in the second picture, yet switching the cup and pen holder’s positions this time.'' \\
    & \textbf{Goal Type}: Image + Language \\
    & \textbf{Description:} The planner must precisely identify the arrangement of objects in the goal image, while also considering the instruction to switch the positions of the cup and the pen holder. \\

    \bottomrule
    
    \end{tabular}
    \caption{\textbf{A list of 4 tasks featuring multimodal goals.} For every task, we provide the instruction as used in our experiments, the goal type, and a detailed description of the task.}
    \label{tab:goal_tasks}
\end{table*}

\begin{table*}[h]
    \small
    \centering
    \begin{tabular}{m{14em}m{40em}}

    \toprule

    \texttt{Stack Blocks} & \textbf{Instruction:} ``Stack all the blocks.'' \\
    & \textbf{Description:}  In this task, we inject noise during the execution of the primitive skill ``Place a block on another block''. In 4 out of 10 evaluation episodes, the primitive skill fails when stacking blocks for the first time. In another 4 episodes, the failure occurs during the second stacking attempt. For the remaining 2 episodes, the primitive skill does not fail. \\

        \noalign{\vskip 1mm}
    \hline
    \noalign{\vskip 1mm}
    \texttt{Pack Chip Bags} & \textbf{Instruction:} ``Put the chip bag on the table in the gift box.'' \\
    & \textbf{Description:} This task involves human intervention, where a person removes the chip bag placed in the gift box by a robot arm. In five of the evaluation episodes, the chip bag that is placed in the gift box for the first time is removed and placed on the table by human. In the other five episodes, the chip bag that is placed in the gift box for the second time is removed and placed on the table. \\

    \noalign{\vskip 1mm}
    \hline
    \noalign{\vskip 1mm}
    \texttt{Find Stapler} & \textbf{Instruction:} ``Put the stapler on the table.'' \\
    & \textbf{Description:} In this task, the target object (stapler) may be placed in the top drawer, bottom drawer, or cabinet. The planner is required to locate the target object based on visual feedback. In three evaluation episodes, the target object is placed in the top drawer; in four episodes, it is placed in the bottom drawer; and in the remaining three episodes, it is placed inside the cabinet. \\

        \noalign{\vskip 1mm}
    \hline
    \noalign{\vskip 1mm}
    \texttt{Human-Robot Interaction} & \textbf{Instruction:} ``Pass me a can of cola.'' \\
    & \textbf{Description:} In this task, the robot arm can only execute ``Place can of cola in human hand'' after detecting a human hand. Before that, the robot repeatedly waits and checks every five seconds for the hand's appearance. In two of the evaluation episodes, the human hand appears directly in the observation; in the remaining eight episodes, the human hand appears in the observation several seconds after the robot arm picks up the can of cola. \\

    \bottomrule
    
    \end{tabular}
    \caption{\textbf{A list of 4 tasks requiring visual feedback.} For every task, we provide the instruction as used in our experiments and a detailed description of the task.}
    \label{tab:feedback_tasks}
\end{table*}

\clearpage
\subsection{Simulated Environment}
\label{sec:app-simulation}

\subsubsection{Tasks}
Drawing inspiration from the Grounded Decoding~\cite{huang2023grounded} setting, we develop 16 tasks based on the RAVENS environment~\cite{zeng2021transporter} (listed in Table~\ref{table:ravens_full}). 
Each task requires a UR5 robot to rearrange the objects on the table in some desired configuration, specified by high-level language instructions. 
A camera is employed to capture a top-down view for task planning purposes.
The tasks are categorized into two categories: (i) \texttt{Blocks \& Bowls} (8 tasks), which focus on rearranging or combining blocks and bowls. (ii) \texttt{Letters} (8 tasks), which involve rearranging alphabetical letters. 
Upon each reset, task-relevant objects, along with some distractor objects, are randomly distributed across the workspace. We employ a binary reward function using ground-truth state of the objects to facilitate automatic evaluations. Additionally, for certain tasks, the attributes mentioned in the instructions are also randomized, details of which are provided below:

\begin{itemize}
    \item \textbf{corner/side}: top left corner, top side, top right corner, left side, right side, bottom right corner, bottom side, bottom left corner
    \item \textbf{word}: cat, dog, red, blue, pink, gold, yoga, fork, soap, milk, dance, bread, knife, chair, peach, white, brown, plate, brush, table
\end{itemize}

\subsubsection{Prompts}
\begin{figure*}[t!]
    \centering
    \includegraphics[width=0.87\linewidth]{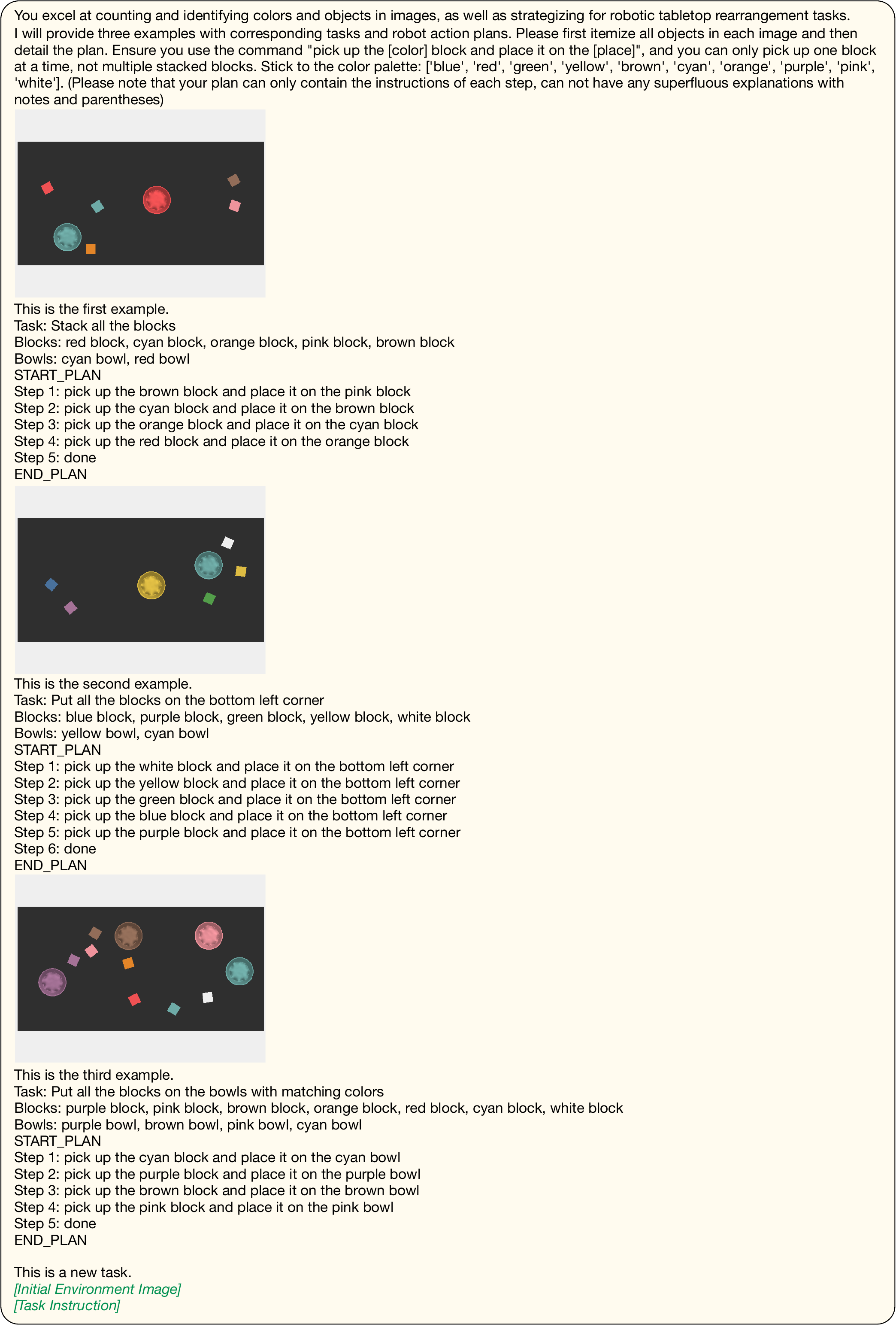}
    \caption{\textbf{Prompt for \texttt{Blocks \& Bowls} in simulated environment.}}
    \label{fig:blocks_and_bowls_prompt}
\end{figure*}

\begin{figure*}[t!]
    \centering
    \includegraphics[width=1.0\linewidth]{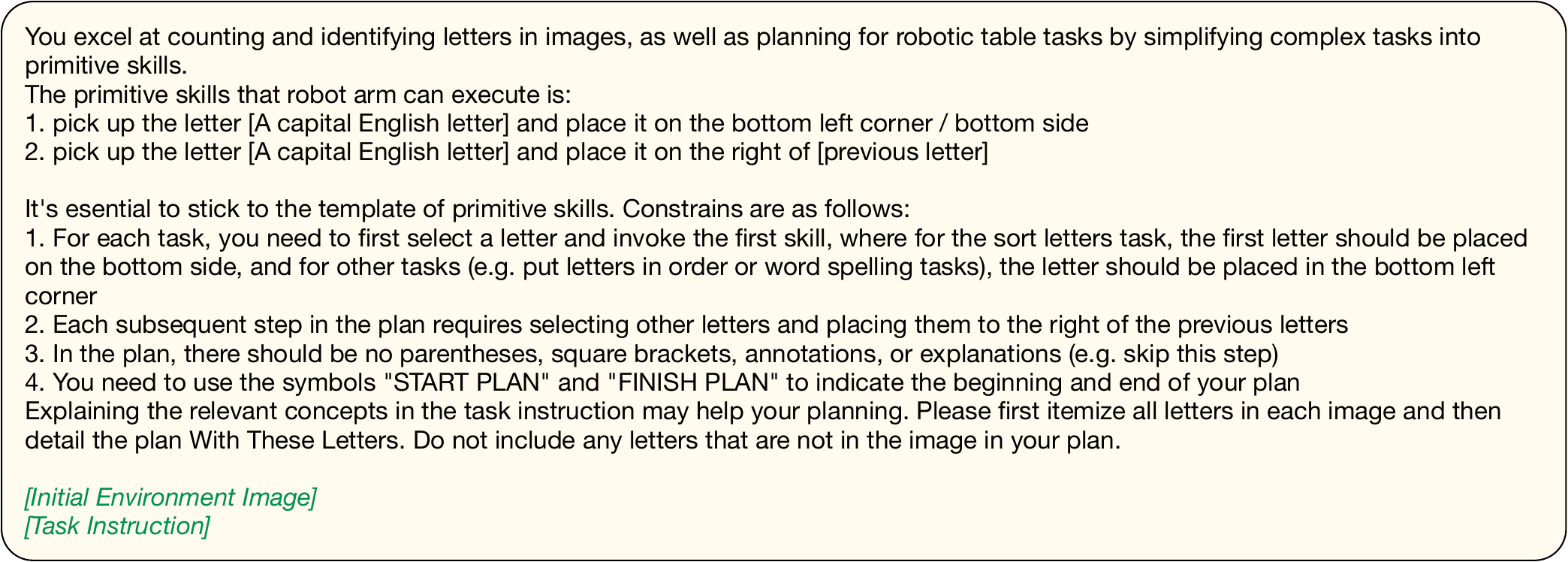}
    \caption{\textbf{Prompt for \texttt{Letters} in simulated environment.}}
    \label{fig:letters_prompt}
\end{figure*}

The prompt for \texttt{Blocks \& Bowls} is shown in Figure~\ref{fig:blocks_and_bowls_prompt}. We incorporate three in-context examples (seen tasks) into the prompt. This approach addresses the substantial domain gap between simulated images of blocks and bowls and their real-world counterparts. Due to this gap, GPT-4V is unable to recognize and comprehend these objects in a zero-shot setting. 
Conversely, for \texttt{Letters}, we omit in-context examples from the prompt (see Figure~\ref{fig:letters_prompt}), as GPT-4V demonstrates proficient recognition and understanding of all letters.

\subsubsection{Primitive Skills}
In prior work, such as Grounded Decoding, CLIPort policies are utilized as low-level primitive skills. However, our findings suggest that this approach does not accurately represent the capabilities of the high-level planner. We observe that, in many tests, CLIPort policies correctly interact with objects even when the planner generate an \textit{incorrect} step. To address this, we shift to employing script policies as our primitive skills. These policies can directly access the ground-truth states of objects within the simulator, ensuring a noise-free outcome and providing a more accurate measure of the planner's success rate.

\subsubsection{Baselines}

Our comparison encompasses three baseline categories: 
\textit{(i)} CLIPort~\cite{shridhar2021cliport}, a language-conditioned imitation learning agent that directly take in the high-level language instructions without a planner. We consider two variants: ``Short'', trained on single-step pick-and-place instructions, and ``Long'', trained on high-level instructions. During evaluation, both CLIPort (Short) and CLIPort (Long) receive only the high-level instructions. These baselines aim to evaluate whether solitary language-conditioned policies can perform well on long-horizon tasks and generalize to new task instructions. Our implementation adheres closely to the description in the Grounded Decoding paper~\cite{huang2023grounded}; for more details, please refer to this paper. 
\textit{(ii)} An LLM-based planner that does not relay on any grounding/affordance model. We evaluate Llama 2 70B and GPT-4. Our evaluation includes Llama 2 and GPT-4. The inclusion of Llama 2 stems from its use in our reimplementation of Grounded Decoding. Grounded Decoding requires access to the output token probabilities from LLMs. However, with the OpenAI API not providing these probabilities, we are constrained to using the open-source Llama 2.
\textit{(iii)} Grounded Decoding (GD), which integrates an LLM with an affordance model for enhanced planning. Here, Llama 2 is used as the LLM. For the \texttt{Blocks \& Bowls} scenario, affordances are deduced from CLIPort's predicted logits, as outlined in the GD paper. For \texttt{Letters}, we resort to ground-truth affordance values from simulation due to the limited generalization capability of CLIPort's predicted logits on unseen letters. We employ the beam search variant of GD.

\subsubsection{Full Results on Simulated Environments}
\label{sec:sim_full_results}
In Table~\ref{table:ravens_full}, we show the full list of tasks in simulated environment, alongside their corresponding experimental results. The tasks are categorized by background color: those with a blue background are `seen' tasks, while those with an orange background are `unseen' tasks. `Seen' tasks are used for training for supervised baselines (CLIPort), or included in prompts for high-level planners. However, in the case of the \model' prompt within the \texttt{Letters} category, we do not include any `seen' tasks.

\begin{table*}[h]
\begin{center}
\setlength\tabcolsep{2.0pt}
\small
\bgroup
\def\arraystretch{1.4}
\begin{tabular}{@{}lcccccc@{}}
    \toprule
      \multicolumn{1}{l}{} & \multicolumn{2}{c}{\textbf{CLIPort}} & \multicolumn{2}{c}{\textbf{LLM}} & \multirow{2}{*}{\textbf{GD}}  & \multirow{2}{*}{\textbf{\model}} \\
      \cmidrule(lr){2-3} \cmidrule(lr){4-5}
    Tasks & Short & Long & Llama 2 & GPT-4  \\

    \midrule
    \normalsize{\textbf{\textcolor{gray}{Blocks \& Bowls}}}\\
    \rowcolor{lightblue}
    \small{Stack all the blocks}   & 10\% & 90\% & 0\% & 0\% & 30\% & 90\% \\
    \rowcolor{lightblue}
    \small{Put all the blocks on the \textit{corner/side}}   & 0\% & 65\% & 0\% & 0\% & 10\% & 90\% \\
    \rowcolor{lightblue}
    \small{Put all the blocks in the bowls with matching colors}   & 0\% & 50\% & 5\% & 0\% & 15\% & 55\% \\
    \rowcolor{lightorange}
    \small{Put the blocks in the bowls with mismatched colors}   & 10\% & 25\% & 0\% & 0\% & 0\% & 80\%\\
    \rowcolor{lightorange}
    \small{Put all the blocks on different corners}   & 0\% & 0\% & 0\% & 0\% & 20\% & 90\% \\
    \rowcolor{lightorange}
    \small{Stack only the blocks of cool colors}   & 5\% & 0\% & 0\% & 20\% & 5\% & 60\% \\
    \rowcolor{lightorange}
    \small{Stack only the blocks of warm colors}   & 15\% & 5\% & 20\% & 10\% & 15\% & 80\% \\
    \rowcolor{lightorange}
    \small{Stack only the primary color blocks on the left side}  & 0\% & 0\% & 80\% & 80\% & 75\% & 95\% \\
    \textbf{Total} & 5.0\% & 29.4\% & 13.1\% & 13.8\% & 21.3\% & 80.0\%\\
    
    \midrule
    \normalsize{\textbf{\textcolor{gray}{Letters}}} \\
    \rowcolor{lightblue}
    \small{Put the letters on the tables in alphabetical order}   & 0\% & 30\% & 0\% & 0\% & 25\% & 95\% \\
    \rowcolor{lightblue}
    \small{Spell as much of \textit{word} as you can}   & 0\% & 55\% & 55\% & 60\% & 80\% & 75\% \\
    \rowcolor{lightblue}
    \small{Sort the vowels from the remaining letters to the bottom side}   & 0\% & 35\% & 20\% & 15\%  & 50\% & 95\% \\
    \rowcolor{lightorange}
    \small{Put the letters on the tables in reverse alphabetical order}   & 0\% & 0\% & 0\% & 0\% & 10\% & 95\% \\
    \rowcolor{lightorange}
    \small{Correctly spell out a sport using the present letters}   & 0\% & 0\% & 0\% & 0\% & 35\% & 85\% \\
    \rowcolor{lightorange}
    \small{Sort the geometrically vertically symmetrical letters to the bottom side}   & 0\% & 0\% & 0\% & 0\% & 55\% & 40\% \\
    \rowcolor{lightorange}
    \small{Sort the consonants from the remaining letters to the bottom side}   & 0\% & 0\% & 0\% & 0\% & 10\% & 90\% \\
    \rowcolor{lightorange}
    \small{Sort the letters less than ``D'' according to ASCII to the bottom side}   & 5\% & 0\% & 75\% & 75\%  & 100\% & 100\% \\
    \textbf{Total}  & 0.6\% & 15.0\% & 18.8\% & 18.8\% & 45.6\% & 84.4\% \\
    \bottomrule
\end{tabular}
\egroup
\caption{\textbf{Full experimental results in simulation on \textcolor{darkblue}{seen tasks} and \textcolor{darkorange}{unseen tasks}.} Each entry represents success rate averaged across 20 episodes.}
\label{table:ravens_full}
\end{center}
\end{table*}